\documentclass[runningheads]{llncs}

\usepackage{eccv}

\usepackage{eccvabbrv}
\usepackage{multirow}
\usepackage{multicol}
\usepackage{graphicx}
\usepackage{booktabs}

\usepackage[accsupp]{axessibility}  

\usepackage{hyperref}

\usepackage{orcidlink}

\begin{document}

\title{Segment, Lift and Fit: Automatic 3D Shape Labeling from 2D Prompts} 


\author{Jianhao Li\inst{1}\and
Tianyu Sun\inst{2} \and
Zhongdao Wang\inst{3}$^\dagger$ \and 
Enze Xie \inst{3}\and
Bailan Feng\inst{3} \and
Hongbo Zhang\inst{3} \and
Ze Yuan\inst{1} \and
Ke Xu \inst{1} \and
Jiaheng Liu \inst{4}$^\dagger$\and
Ping Luo \inst{5}
}

\authorrunning{J. Li et al.}

\institute{Beihang University, Beijing, China \and
Tsinghua University, Beijing, China \and
Noah's Ark Lab, Huawei \and
Nanjing University \and
The University of Hong Kong\\
\email{\{lijianhao, yuanze1024, kexu\}@buaa.edu.cn\\
\{wangzhongdao, Johnny\_ez, fengbailan, zhanghongbo888\}@huawei.com\\
sty21@mails.tsinghua.edu.cn, buaaljiaheng@gmail.com, pluo.lhi@gmail.com
}
\footnotetext{$^\dagger$ Corresponding authors: \email{Zhongdao Wang; Jiaheng Liu}.}
}

\maketitle

\begin{abstract}
  This paper proposes an algorithm for automatically labeling 3D objects from 2D point or box prompts, especially focusing on applications in autonomous driving. Unlike previous arts, our auto-labeler predicts 3D shapes instead of bounding boxes and does not require training on a specific dataset.
We propose a \emph{Segment, Lift, and Fit} (SLF) paradigm to achieve this goal. Firstly, we \emph{segment} high-quality instance masks from the prompts using the Segment Anything Model ({SAM}) and transform the remaining problem into predicting 3D shapes from given 2D masks.
Due to the ill-posed nature of this problem, it presents a significant challenge as multiple 3D shapes can project into an identical mask.
To tackle this issue, we then \emph{lift} 2D masks to 3D forms and employ gradient descent to adjust their poses and shapes until the projections \emph{fit} the masks and the surfaces conform to surrounding LiDAR points. Notably, since we do not train on a specific dataset,  the SLF auto-labeler does not overfit to biased annotation patterns in the training set as other methods do. Thus, the generalization ability across different datasets improves. Experimental results on the KITTI dataset demonstrate that the SLF auto-labeler produces high-quality bounding box annotations, achieving an AP@0.5 IoU of nearly 90\%. Detectors trained with the generated pseudo-labels perform nearly as well as those trained with actual ground-truth annotations. Furthermore, the SLF auto-labeler shows promising results in detailed shape predictions, providing a potential alternative for the occupancy annotation of dynamic objects.
\end{abstract}

\section{Introduction}
\label{sec:intro}

\begin{figure*}[t]
    \centering
    \includegraphics[width=1.0\linewidth]{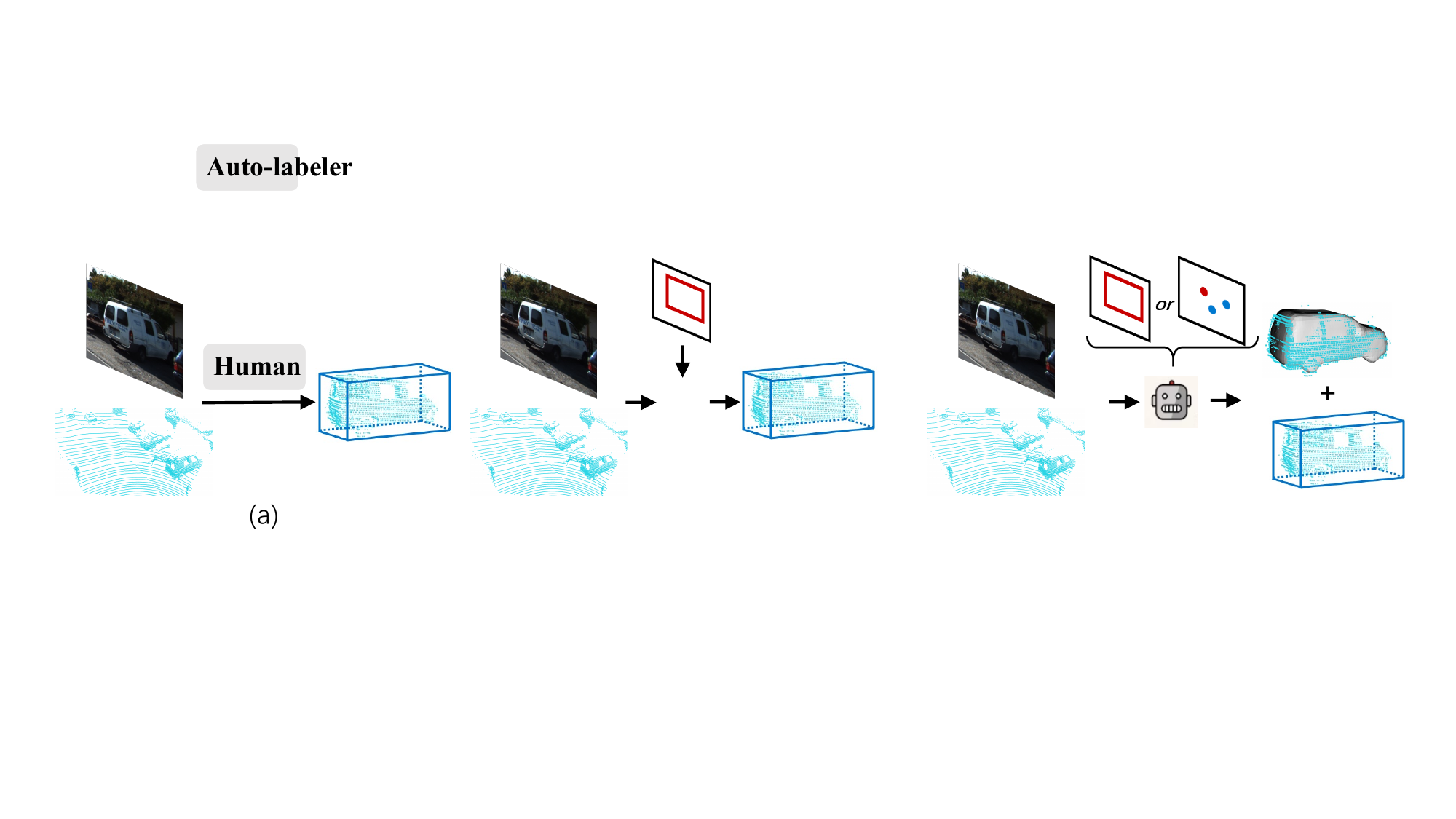}
    \caption{Comparison between human annotation and auto-labeler annotation using SLF.}
    \label{fig:intro}
\end{figure*}

In modern robotics and autonomous driving, a significant amount of labeled data is required for supervised learning to understand the 3D scene~\cite{liu2021geometrymotion,liu2022apsnet,guo20223d,liu2022geometrymotion,Liu_2024_CVPR}, emphasizing dynamic objects such as vehicles and pedestrians. Therefore, there is a pressing need to improve the degree of automation for 3D labeling. The preferred labeling format is typically the 3D bounding box, which can be represented compactly with as few as 7-9 degrees of freedom. However, manually labeling a large number of 3D boxes remains a laborious and costly task, posing challenges for scaling up 3D object detectors as shown in Fig.~\ref{fig:intro}(a).

Meanwhile, as 3D perception models evolve, there is an increasing demand for more precise labeling granularity. While bounding boxes offer a compact representation, they are not optimal for 3D objects and are being replaced by finer representations. For instance, voxel occupancy ~\cite{surrdocc,tpvformer,openocc} has emerged as a dominant format for 3D scenes in autonomous driving, as it unifies the representation of static road elements and dynamic objects. However, finer-grained labels further complicate the annotation pipeline, exacerbating the annotation efficiency problem and presenting a significant obstacle to scaling up 3D models.

In this paper, we propose a novel approach for 3D auto-labeling. The proposed auto-labeler follows a \emph{Segment, Lift, and Fit} (SLF) paradigm as shown in Fig.~\ref{fig:intro}(b).
Specifically, SLF utilizes the 2D points and bounding boxes as prompts and outputs the estimated 3D shapes and poses of target objects, thus supporting semi- or fully automatic labeling depending on whether the prompts come from humans or machines. To accomplish this, we leverage the {Segment Anything Model (SAM)}~\cite{sam}, a foundational vision model, to generate high-quality instance masks from the input prompts. Subsequently, we lift 2D instance masks into 3D forms with Signed Distance Functions (SDFs)~\cite{sdf}, and then iteratively optimize the shapes and poses until their projections fit the mask and its surface conforms to the surrounding LiDAR points. Notably, the proposed SLF auto-labeler does not require supervised training on a specific dataset, and is therefore not susceptible to overfitting biased annotation patterns in the training dataset.

Experimental results show that SLF can generate high-quality 3D labels. As our method outputs detailed shapes of objects, while existing benchmarks~\cite{kitti,nuscenes} only provide bounding box annotations,
for a fair comparison with existing auto-labelers, we convert shapes to boxes and assess the label quality regarding box accuracy. When compared among unsupervised auto-labelers~\cite{sdflabel,fgr,liftanddiscount} that do not train with 3D labels as ours, our method outperforms the best-performing FGR~\cite{fgr} auto-labeler in both settings of direct evaluation on pseudo labels and training object detectors with generated pseudo labels. Moreover, because SLF does not learn biased annotation patterns from the training data, it significantly outperforms state of art supervised model MTrans~\cite{mtrans} on “unseen” datasets, demonstrating its superior generalization capacity. We also show qualitatively that the predicted shapes align well with the actual geometry of the target objects, underpinning a new possibility for voxel occupancy annotation~\cite{surrdocc,openocc}.

Our contribution can be summarized as follows:
\begin{itemize}
\item We propose SLF, a 3D auto-labeler that improves existing methods by predicting detailed shapes and poses rather than just bounding boxes. The shape and pose are directly optimized using gradient descent without requiring a neural network trained on a small portion of labels, which enables a “cold-start” scenario for auto-labeling.
\item We demonstrate that SLF generates superior labels to other unsupervised auto-labelers. Additionally, it shows good generalization ability across different datasets, surpassing supervised auto-labelers. 

\item The predicted shapes align well with the actual geometries of the objects, suggesting that SLF can serve as a viable alternative solution for annotating fine-grained occupancy of dynamic objects.
\end{itemize}

\section{Related Works}
\noindent\textbf{3D Auto-labelers.}
Compared to supervised 3D object detectors~\cite{pointpillar,second,liu20223d,yang2023gd}, auto-labelers require fewer or no training labels.
\emph{Supervised} auto-labelers learn to annotate from limited samples with ground-truth 3D labels, often in the hundreds. For instance, WS3D~\cite{ws3d} generates cylindrical object proposals on bird's eye view (BEV) maps using center clicks in BEV and fine-tunes a 3D detector pre-trained on a few human-annotated data. MTrans~\cite{mtrans} employs a multi-task design of segmentation, point generation, and box regression to train a 3D detector that takes ground truth 2D boxes and LiDAR points as input.
\emph{Unsupervised} auto-labelers do not require 3D labels for training. CC~\cite{cc} uses PointNet~\cite{pointnet} to segment and regress 3D boxes from frustum points. SDFLabel~\cite{sdflabel} trains a DeepSDF~\cite{deepsdf} network supervised with image patch and  CAD models of cars on a synthetic dataset, then uses RANSAC~\cite{ransac} to fit the predicted shape to frustum sub-points. VS3D~\cite{vs3d} employs an unsupervised UPM module to generate 3D proposals and filter them using a pre-trained image model.
Our SLF, is also an unsupervised approach that does not require training with 3D labels. SLF achieves finer annotations with weaker prompts compared with a closely related unsupervised and non-learning approach, FGR~\cite{fgr}. FGR requires \emph{amodal} boxes as inputs, in which oracle occlusion information has already been considered. LPCG~\cite{peng2022lidar} finds that concerning label accuracy, the 3D location part in the label is preferred compared to other parts of labels in monocular 3D object detection and proposed training-free and high-cost label pipeline, which is similar to FGR and MTrans respectively. In contrast, SLF takes normal 2D \emph{modal} bounding boxes as input, which an off-the-shelf 2D detector can easily obtain. Furthermore, SLF can produce 3D object shapes rather than just bounding boxes.


\noindent\textbf{Shape Prior.}
To reconstruct the 3D geometry for instances from a specific semantic class, \eg, cars or pedestrians, the shape prior of the given class is often utilized to ease the estimation.
For instance, McCraith et al.~\cite{liftanddiscount} uses a fixed car shape as the most straightforward shape prior. 
Autoshape~\cite{autoshape} develops a shape-labeler for the car class, but it requires ground truth 3D bounding box and amodal instance mask which contains oracle occlusion information as input and it uses part-based deformable representation as shape prior which needs to label the key points and different part of CAD model.
Another usage of shape prior is to encode it into a compact low-dimensional space with PCA~\cite{jointopt,directshape,simmono,samp} or neural network~\cite{sdflabel,3d-rcnn}. 
SDFLabel~\cite{sdflabel} pretrains a DeepSDF~\cite{deepsdf} model to predict the hidden representation of shape conditioned on cropped image patches. 3D-RCNN~\cite{3d-rcnn} regresses poses and shape parameters in a single forward pass. However, in 3D-RCNN shape representation is not differentiable, and the finite differences that are used for gradient approximation cause ambiguity during training. Upon the compact PCA representation, Engelmann et al.~\cite{jointopt} and DirectShape~\cite{directshape}  optimize the shape and pose concurrently with the stereo reconstruction of the object to refine the 3D detection result. SAMP~\cite{samp} takes sequence of stereo depth maps and 3D trajectory to generate target's shape and refine its trajectory. Prisacariu et al.~\cite{simmono} requires sequence of instance masks and results of 3D tracker to optimize 3D model. Similar to previous PCA based method, our method adopts the PCA shape prior in a differentiable optimization process, however, SLF is training-free and only requires a single raw point cloud and RGB image with 2D prompts, which is more practical.

\section{Approach}
\begin{figure*}[t]
    \centering
    \begin{minipage}[t]{\linewidth}
        \centering
         \begin{subfigure}{0.32\linewidth}
		\includegraphics[width=\textwidth,height=1\textwidth]{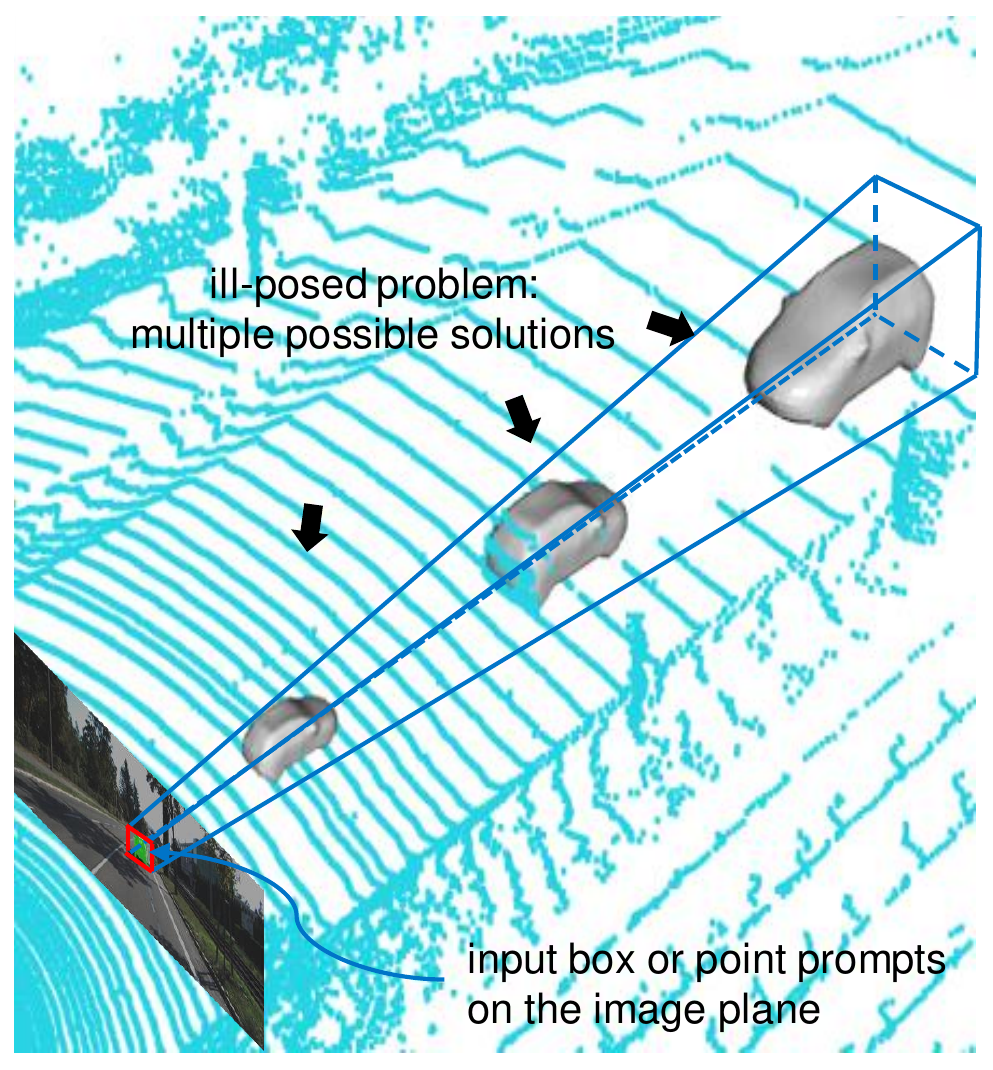}
		\caption{Definition \& challenge}
	\end{subfigure}
    \hfill
    \begin{subfigure}{0.67\linewidth}
		\includegraphics[width=\textwidth,height=0.48\textwidth]{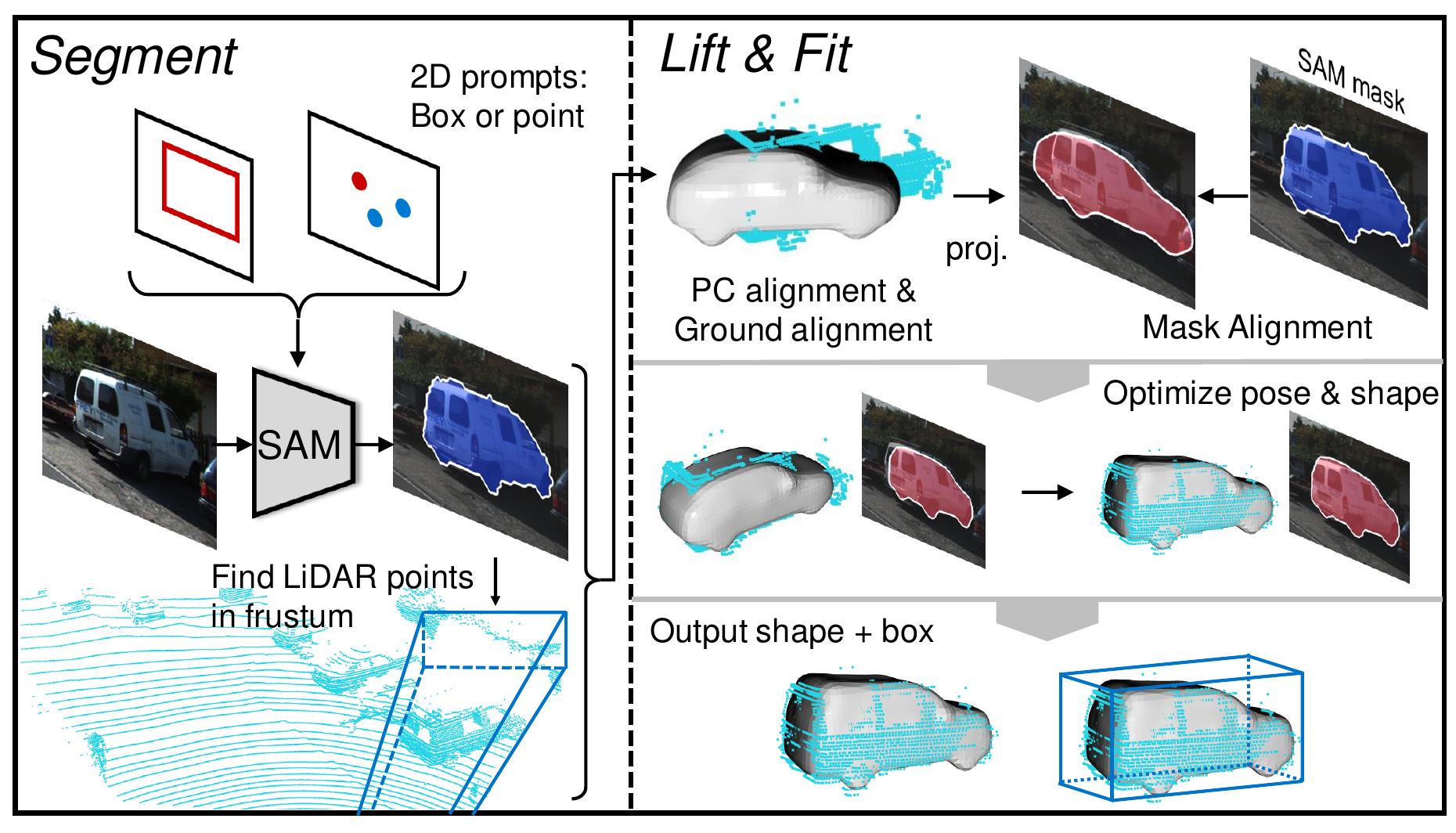}
		\caption{Overall illustration of the proposed method}
	\end{subfigure}
    \end{minipage}

    \caption{ (a) We aim to recover the 3D shape and pose of an interested object, given 2D box or point as 
    prompts. This problem is highly ill-posed. (b) In our SLF, we propose to firstly \emph{segment} the 2D mask of the target object, and then \emph{lift} the mask to a 3D form and optimize over its shape and pose by gradient descent until the 3D object \emph{fits} the mask and LiDAR points.}
    \label{fig:brief_illustration}
\end{figure*}

As shown in \cref{fig:brief_illustration}(a),
we aim to recover the 3D shape and pose of an interested object by asking a labeler to click one or more points or draw a box on the image as the prompt for the target object. 
A {labeler} could be a human in semi-automatic labeling and a machine algorithm such as a 2D object detector in case of full-automatic labeling.
\cref{fig:brief_illustration}(a) shows a brief illustration.  
We propose a \textit{Segment, Lift, and Fit} (SLF) paradigm to address this problem.
We first segment a 2D mask of the target object using the input prompts, then lift it into a 3D model and iteratively optimize its shape and pose until it fits the 2D mask and surrounding LiDAR points. 
An overall illustration is shown in \cref{fig:brief_illustration}(b).

\subsection{Segment: From 2D Prompts to Instance Masks}
\label{sec:def}
The problem of auto-labeling 3D shapes from 2D prompts is a significant issue in autonomous driving because it would be highly beneficial to effectively utilize the already established robust 2D instance segmentation models for this purpose.
This work employs the strong Segment Anything Model (SAM)~\cite{sam} to transfer the input point or box prompts into 2D instance masks. 
Then, our problem is transferred to estimate the 3D shapes and poses from given instance masks. 
Despite the high quality of SAM masks, the problem is still extremely difficult to solve. This is because multiple 3D shapes can be mapped to the same 2D mask (\cref{fig:brief_illustration}(a)), making it a highly ill-posed problem and posing a significant challenge for auto-labeler methods. 

\subsection{Optimization Objective}
We employ a \emph{pose} vector $\mathbf{p}=(x,y,z,\theta)\in \mathbb{R}^4$ and a latent \emph{shape} vector $\mathbf{s} \in \mathbb{R}^d$ to represent a 3D object, where $(x,y,z)$ denote the position of the object center in the global coordinate system, $\theta$ denotes the heading angle around the $z$ axis, and the latent dimension $d$ is empirically set to $5$. We assume all instances are on the ground (the $x$-$y$ plane) and omit rotations around the $x$ and $y$ axes. Our goal is to minimize an energy function that corresponds to the posterior probability of the 3D reconstruction given $\mathbf{p}$ and $\mathbf{s}$ as follows:
\begin{equation}
\label{eq:overall}
    \mathbf{p}^{*}, \mathbf{s}^{*} = \arg\min_{\mathbf{p}, \mathbf{s}} E(Y,\mathcal{X}_f,\mathbf{p}, \mathbf{s}),
\end{equation}
where $Y\in [0,1]^{h\times w}$ is the 2D instance mask predicted by SAM, 
and $\mathcal{X}_f \in \mathbb{R}^{N\times 3}$ is a subset of the whole scene point cloud $\mathcal{X}$. 
$\mathcal{X}_f$ is comprised of $N$ three-dimensional coordinates of points inside the camera frustum corresponding to the 2D mask. 
To solve the optimization problem in \cref{eq:overall}, 
we build a differentiable renderer to optimize the energy function $E$ via gradient descent. 

\begin{figure*}[t]
    \centering
    \includegraphics[width=\linewidth,height=0.26\textwidth]{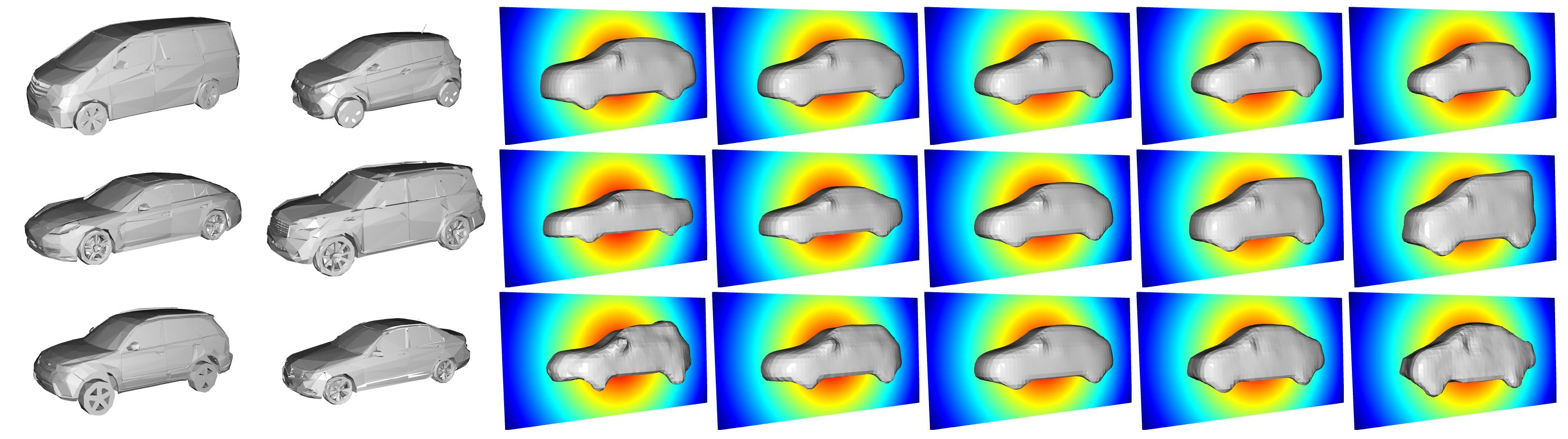}
    \caption{We perform PCA on a diverse collection of 3D shapes to obtain the shape latent code. \textbf{Left}: Example of the basis 3D shapes in the collection. \textbf{Right}: Interpolation along the latent space of the shape code (each row)  shows smooth shape variations.}
    \label{fig:shape_prior}
\end{figure*}
\subsection{Lift: Differentiable Rendering }
\label{sec:lift}

\noindent\textbf{Representation of 3D objects.} 
We use the Signed Distance Function (SDF)~\cite{directshape,autoshape} to represent 3D objects. 
Formally, an SDF is a three-dimensional scalar field $\phi(\mathbf{x})$ in which the value of a particular spatial coordinate $\mathbf{x}$ is given by its distance to the closest object surface, with positive and negative values meaning exterior and interior area. 
In practice, we use the discrete approximation of continuous SDF.
Consider a cubic divided into a 3D mesh grid with the size of $l \times w \times h$. We can represent the inscribed object with the SDF value in each mesh grid, \ie, $\mathbf{m} = \left[f(\mathbf{x}_1),...,f(\mathbf{x}_i)...,f(\mathbf{x}_{n}) \right]^{T}$, where $n=l \times w \times h$.

\noindent\textbf{Shape prior.} 
We aim to condense the nuanced differences present in class-specific 3D shapes into a low-dimensional latent space, where traversing along the manifold of the latent vector $\mathbf{s}$ corresponds to the exploration of feasible shapes of the object. We focus on the ``car'' class in this paper, where every instance in the class possesses an overall shape prior and exhibits minor variations in characteristics, such as the shape of the front face or the shape of the cabin.

To obtain the shape prior, we first assemble a collection of 3D vehicle models from the Apollo-Car3D dataset~\cite{apollo}, which consists of 79 models with exhibit significant variations in shapes and sizes (see the left of \cref{fig:shape_prior}). 
However, using the Signed Distance Function (SDF) representation necessitates that objects be water-tight to avoid ambiguity between their interior and exterior. Some models contain intricate internal structures and are not initially water-tight. 
To obtain the water-tight surface of 3D models, we employ the SoftRas~\cite{softras} algorithm to remove redundant vertices and faces inside the model. 
Finally, we convert the original triangular mesh representation of each 3D model into an SDF representation to obtain a set of models $\mathcal{M} = \{\mathbf{m}_{i}\}_{i=1}^{79}$.

Given the collection of representative 3D models, we perform the PCA method on $\mathcal{M}$,
and keep the top-$d$ ($d=5$) principle components to obtain a projection matrix $V\in \mathbb{R}^{d \times n}$. Accordingly, we can use the projection matrix $V$ to render a 3D shape $\mathbf{m}$ from a given shape vector $\mathbf{s}$. Besides, we can encode a 3D model $\mathbf{m}$ into a shape vector $\mathbf{s}$ by
$\mathbf{s}=V^\top(\mathbf{m}-\mathbf{\bar{m}}) $ and 
$\mathbf{m}=V\mathbf{s}+\mathbf{\bar{m}}$,
where $\mathbf{\bar{m}}$ indicates the mean vector of the collection  $\mathcal{M}$. In this work, we select a compact representation $d=5$.  The right of \cref{fig:shape_prior} illustrates that interpolation along the latent space of $\mathbf{s}$ leads to smooth variation in shapes. 

\noindent\textbf{Differentiable rendering.} After recovering the 3D shape in a local coordinate system, we can put the model in the global coordinate system by adding its pose $\mathbf{p}$.
This gives us an unbounded SDF $\phi (\textbf{x})$ of the 3D model, in which the values outside the pre-defined cubic are filled with negative infinity. Notably, the rendering process of shape is differentiable \wrt both $\mathbf{s}$ and $\mathbf{p}$.

\subsection{Fit: Optimization over Pose and Shape}
\label{sec:fit}

\begin{figure}[t]
    \centering
    \begin{subfigure}{0.23\linewidth}
		\includegraphics[width=\textwidth]{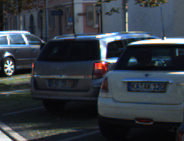}
		\caption{RGB image}
	\end{subfigure}
    \begin{subfigure}{0.23\linewidth}
		\includegraphics[width=\textwidth]{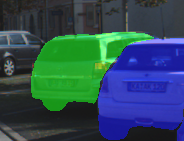}
		\caption{SAM mask}
	\end{subfigure}
     \begin{subfigure}{0.23\linewidth}
            \includegraphics[width=\textwidth]{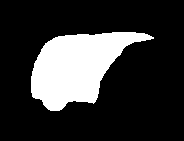}
            \caption{Target mask}
    \end{subfigure}
     \begin{subfigure}{0.23\linewidth}
            \includegraphics[width=\textwidth]{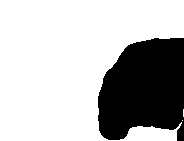}
            \caption{Occ. mask}
    \end{subfigure}

\caption{Generation of occlusion map $O$ for the mask alignment objective.}
\label{fig:maskloss}
\end{figure}

Given the 2D instance mask and the point cloud scene, we randomly initialize pose $\mathbf{p}$ and shape $\mathbf{s}$ and optimize them using \cref{eq:overall}. 
Specifically, the optimization objective consists of three terms: the mask alignment term, the point cloud alignment term, and the ground alignment term. 

\noindent\textbf{Mask alignment term.} The mask alignment term measures the dissimilarity between the target instance segmentation mask $Y$ and the mask $Y_{{proj}}$ obtained by projecting the 3D model onto the image plane. To make the projection process differentiable, we follow~\cite{autoshape,directshape} to define the projection as  
$\pi (p_j)=1-\prod_{\mathbf{x} \in \mathcal{R}_{p_j}} \frac{1}{\mathrm{e}^{\phi(\mathbf{x}) \zeta }+1} $,
where $p_j$ is the $j$-th pixel in the image plane and $\pi(p_j) \in (0,1)$ is the mask value. A ray is cast through $p_j$ from the ego camera, and a set of points $\mathcal{R}{p_j}$ is collected by sampling along the ray. If the ray does not hit the surface of the 3D model, the SDF values of all sampled points $\mathbf{x} \in \mathcal{R}{p_j}$ will be negative, resulting in a mask value close to zero. Conversely, if the ray intersects with the 3D model, there must exist sampled points with $\phi(\mathbf{x}) > 0$, leading to a mask value close to 1. A hyperparameter $\zeta$ controls the ``hardness'' of the mask values,
where a larger $\zeta$ leads to harder values (close to either 0 or 1).
Finally, we define the mask alignment term using the dice loss~\cite{dice} as follows:
\begin{equation}
\label{eq:mask}
    E_{mask}=1-\frac{2\left\lvert (Y_{proj}*O)\bigcap Y\right\rvert }{\left\lvert Y_{proj}*O\right\rvert +\left\lvert Y\right\rvert },
\end{equation}
where $O$ is a binary map indicating if the pixel is occluded by a nearer object. 
As shown in \cref{fig:maskloss},
to obtain the occlusion map, 3D LiDAR points are associated with each 2D mask based on projections, and the depth of each mask is computed as the median depth value of the associated LiDAR points. Masks are sorted by their depth values and then used to compute the occlusion maps. 

\noindent\textbf{Point cloud alignment term.}
Solely optimizing the mask alignment term leads to inaccurate estimation due to the missing depth information. To avoid undesired local optimum, we introduce an additional point cloud alignment term. This term measures how well the 3D object's surface aligns with the point cloud inside the corresponding frustum.

Intuitively,  we can use the $\ell$-1 norm of the SDF $\phi(\mathcal{X}_f)$ to determine how well a 3D model fits a given frustum point cloud, denoted as $\mathcal{X}_f$. However, this term alone might lead to sub-optimal solutions, as the points could be associated with the back face of the object. 
To overcome this issue, a second term is introduced that involves casting a ray for each point $\mathbf{x}\in \mathcal{X}_f$ starting from the ego camera and recording the coordinate $\mathbf{y}$ when the first time the ray intersects with the 3D model. This term penalizes the predicted 3D model if the distance between $\mathbf{x}$ and $\mathbf{y}$ is considerable. The point cloud alignment term can be written as follows:
\begin{equation}
\label{eq:pc}
    E_{pc} =  \Vert \phi(\mathcal{X}_f) \Vert_1 + 
    \frac{1}{\vert \mathcal{\hat{X}}_f \vert }\sum_{\mathbf{x}\in{\mathcal{\hat{X}}_f}}\Vert \mathbf{x} -\mathbf{y} \Vert_2^2.
\end{equation}
 $\mathcal{\hat{X}}_f$ is a subset of  $\subseteq \mathcal{X}_f$ and represents points whose corresponding ray intersects with the model surface. 
 
\noindent\textbf{Ground alignment term.}
Another strong cue that can be leveraged to refine the 3D pose is the height of the ground since we can assume that vehicles run on the road surface. Specifically, we  regress a ground plane function $z = g(x,y)$ using RANSAC~\cite{ransac} and then define a ground alignment objective as follows:
\begin{equation}
    E_{ground} = \Vert \hat{z} - \frac{\hat{h}}{2} - g(\hat{x},\hat{y}) \Vert^2_2,
\end{equation}
where $(\hat{x}, \hat{y},\hat{z})$ is the predicted object center, and $\hat{h}$ is the height of the object inferred from the predicted 3D shape.

\noindent\textbf{Optimization.} 
The optimization objective is formulated as a weighted sum of $E_{mask}$, $E_{pc}$, and $E_{ground}$. We initialize pose $\mathbf{p}$ with the median value of frustum points and shape $\mathbf{s}$ with mean shape $\mathbf{\bar{m}}$. Based on the Adam optimizer~\cite{adam} with a fixed learning rate of $0.1$ and 150 iterations, a mini-batch comprising all instances (typically $<$ 100) in a scene is employed for optimization, as opposed to individual objects. This approach yields a $5\times$ acceleration
with negligible accuracy drop.

\section{Experiments}
\label{sec:experiments}
Previous auto-labelers can be divided into supervised ones and unsupervised ones. 
The proposed SLF is an unsupervised method. Therefore, we primarily compare it with existing unsupervised methods. Nonetheless, we also compare it with supervised methods and demonstrate that SLF exhibits superior transfer ability across datasets.
In addition to bounding box accuracy, we also evaluate the quality of the predicted 3D shape.

\subsection{Comparison with Unsupervised Auto-labelers}
\label{sec:comparison_unsupervised}
\label{sec:automatic_anno_res}
\noindent\textbf{Dataset and metric.} We conduct experiments on the commonly used KITTI~\cite{kitti} dataset. We follow the official training and validation split to get 3712 training frames and 3768 validation frames. To align with prior works\cite{fgr,mtrans}, the “Car” category is the focus of auto-labeling efforts. The Bird's Eye View (BEV) AP and the 3D AP are selected as evaluation metrics for a comprehensive comparison. Results are reported for all three levels of difficulty (Easy, Moderate, and Hard). The commonly used KITTI metrics are evaluated at a strict threshold of 0.7 IoU. As observed in~\cite{sdflabel, liftanddiscount}, the 3D boxes in KITTI have varying spatial padding, making it challenging for human annotators to determine a precise size for a significant number of instances. Therefore, we follow the previous work~\cite{sdflabel,liftanddiscount} and set the IoU threshold as 0.5 and calculate AP by 40 recall positions.

\begin{table}[t]
\caption{Comparison with unsupervised auto-labelers.}
    \centering
    \begin{tabular}{l  c c c c c c }
    
    \toprule
         \multirow{2}{*}{Method}  & \multicolumn{3}{c}{AP$_{BEV}$(IoU=0.5)$|_{R40}$} & \multicolumn{3}{c}{AP$_{3D}$(IoU=0.5)$|_{R40}$} \\
         \cmidrule(lr){2-4} 
         \cmidrule(lr){5-7}
        & Easy & Mod. & Hard & Easy & Mod. & Hard \\
        \midrule
        VS3D~\cite{vs3d}&  74.5 & 66.7  & 57.6 & 40.3 & 37.4 & 31.1 \\ 
        SDFLabel~\cite{sdflabel} & 77.8 & 59.8 & - & 62.3 & 42.2 & - \\
        McCraith et al.~\cite{liftanddiscount}& 75.5 & 76.6  & 68.6 & - &-  & - \\ 
        
        FGR~\cite{fgr}&  \textbf{93.7} & 77.8  & 70.5 & \textbf{91.3} & 77.8 & 70.6 \\ 
        \midrule 
        SLF${_\texttt{GTmask}}$  & 89.2& 88.9 & \textbf{87.1} & 87.6 & 87.8 & 83.8 \\
        SLF${_\texttt{point}}$  & 88.5 & 88.5 & 84.2 & 84.8 & 84.9 & 80.8 \\
        SLF${_\texttt{box}}$  & 89.2& \textbf{89.3} & 84.7 & 87.9 & \textbf{88.2} & \textbf{83.9} \\
    \bottomrule
    
    \end{tabular}
    
    \label{tab:unsupervised}
\end{table}

\noindent\textbf{Direct evaluation on generated pseudo labels.} A straightforward evaluation is to compare the quality of auto-labeled 3D bounding boxes in terms of the canonical metric, Average Precision (AP). However, the AP metric needs a confidence score for each bounding box for thresholding and integration, but FGR~\cite{fgr} and SLF do not output confidence scores. To remedy this issue, we simply define IoU between the predicted mask and the target mask in SLF as a confidence score. As for FGR, we first project a 3D bounding box to image plane to get a 2D box, and the confidence score is defined by IoU between the predicted 2D box and the ground truth 2D box.

\cref{tab:unsupervised} compares the proposed SLF with existing unsupervised methods on KITTI validation set. SLF${_\texttt{point}}$ uses 2D points as input, corresponding to the semi-automatic labeling scenario with humans in the loop. The point prompts are simulated by random sampling 3 points inside the ground truth masks\cite{kittiinsgt}. SLF${_\texttt{box}}$ indicates 2D bounding boxes as input, corresponding to the full-automatic labeling scenario in which 2D detectors generate the prompts. For all methods that accept 2D boxes as inputs, we use ground truth boxes for a fair comparison. 
We also provide a variant of our method in which the SAM masks are replaced with ground truth masks from~\cite{kittiinsgt}, denoted as SLF${_\texttt{GTmask}}$.

The last three rows in \cref{tab:unsupervised} show the results of SLF with different 2D prompts. SLF${_\texttt{point}}$ and SLF${_\texttt{box}}$ perform roughly the same in terms of AP$_{BEV}$, and SLF${_\texttt{box}}$ shows superior AP$_{3D}$ performance. 
To our surprise, SLF${_\texttt{box}}$ is slightly better than SLF${_\texttt{GTmask}}$ in terms of both AP$_{3D}$ and AP$_{BEV}$ except on hard samples, which means the 2D mask generated by SAM from box prompts are even better than those annotated by humans.  This suggests that SLF benefits from more accurate input prompts, and more accurate instance masks.

We compare SLF${_\texttt{box}}$ with other unsupervised methods as they accept the same input prompts. It can be observed that SLF outperforms FGR and SDFLabel by a significant margin on the Moderate and Hard samples. However, SLF performs worse than FGR on the Easy samples, and we assume that
easy cases in KITTI have much denser sub-points, which are beneficial to the segmentation and box estimation stage of FGR. Moreover, FGR consumes ground-truth \textit{amodal} bounding box, while in KITTI these boxes are obtained by projecting 3D ground-truth boxes to 2D images, which may leak the oracle occlusion and size information of partially visible objects.
SDFLabel uses a similar 3D shape representation to SLF but relies on a synthetic dataset for training, which may limit its performance on real-world data. In contrast, SLF does not require synthetic training data, and generalizes well to real-world scenarios.

\noindent \textbf{Detector performance trained with pseudo labels.}
Following previous work~\cite{fgr,mtrans,ws3d}, to further evaluate the quality of generated pseudo labels, we use them as training samples to train off-the-shelf detectors and compare the results of different detectors. We mainly compare with the previous best-performing FGR auto-labeler and train 4 different 3D detectors.
Specifically,
for all detectors, we use the OpenPCDet~\cite{openpcdet} with default hyperparameters.
In \cref{tab:training}, we compare the results of detectors trained on KITTI training set with ground truth (GT) labels, FGR pseudo labels,  and SLF pseudo labels, respectively.  If not specified, SLF refers to SLF${_\texttt{box}}$ in the remaining sections.

\begin{table}[t]
\caption{Detector performance trained with pseudo labels.}

    \centering
    \begin{tabular}{l c c c c c c c }
    \toprule
         \multirow{2}{*}{Detectors} &  \multirow{2}{*}{Labels}  & \multicolumn{3}{c}{AP$_{BEV}$(IoU=0.5)$|_{R40}$} & \multicolumn{3}{c}{AP$_{3D}$(IoU=0.5)$|_{R40}$} \\
         \cmidrule(lr){3-5} 
         \cmidrule(lr){6-8}
        && Easy & Mod. & Hard & Easy & Mod. & Hard \\
        \midrule
        \multirow{3}{*}{PointPillar~\cite{pointpillar}}
        &GT &95.7 &94.3 &93.2 &95.7 &93.6 &91.4 \\
        &FGR&  96.5 & 90.2  & 87.1 & 94.9 & 89.4 & 84.8 \\ 
        &SLF  &95.3  &94.1  &91.1  &95.2  &92.1  &88.8  \\
        \hline
        \multirow{3}{*}{SECOND~\cite{second}}
        &GT &95.9 &94.8 &93.9 &95.8 &94.6 &91.9 \\
        &FGR&  95.4 & 88.9  & 86.2 & 95.3 & 88.6 & 85.7 \\ 
        &SLF  & 95.6 &94.2 &91.7  &95.5  &92.4 &89.3 \\
        \hline
        \multirow{3}{*}{Part-$A^2$~\cite{parta2}}
        &GT &96.0 &94.3 &93.9 &96.0 &94.2 &91.9 \\
        &FGR&  96.1 & 92.1  & 87.5 & 96.0 & 90.0 & 85.1\\ 
        &SLF  & 96.0 &94.5 &91.8 &95.9  &93.9  &89.6 \\
        \hline
        \multirow{3}{*}{Voxel R-CNN~\cite{voxelrcnn}}
        &GT &98.4 &94.9 &94.5 &98.3 &94.8 &94.4 \\
        &FGR&  98.7 & 90.9  & 88.2 & 98.6 & 90.7 & 88.0  \\ 
        &SLF  & 96.2 &95.1 &92.4  &96.1 &94.7  &91.9  \\
    \bottomrule
    \end{tabular}
    
    \label{tab:training}
\end{table}

\cref{tab:training} shows SLF pseudo labels significantly improve upon FGR pseudo labels on all detectors 
on Moderate and Hard samples. 
Regarding Easy samples, SLF pseudo labels are on par with FGR pseudo labels on all detectors except Voxel R-CNN\cite{voxelrcnn}. When comparing specific detectors,  the one trained with SLF labels performs almost as well as the one trained with ground truth labels on both Easy and Moderate samples, and slightly worse on Hard samples, which shows the effectiveness of our method.

\subsection{Cross Dataset Generalization}
\label{sec:comparison_supervised}
Apart from KITTI, we conduct experiments on more challenging  nuScenes~\cite{nuscenes}. 
We transform the validation set of nuScenes to KITTI format, resulting in 6019 frames with corresponding front view images, point cloud and 3D bounding box annotations. 
On  KITTI, we compare the pseudo label with ground truths and report AP$_{3D}$,
and on nuScenes, we report the official metrics including mAP, NDS, mATE, mASE and mAOE.
We compare with the best-performing unsupervised auto-labeler FGR, and the supervised auto-labeler MTrans~\cite{mtrans}. 
Both FGR and Mtrans require 2D amodal boxes as prompts, so we project 3d ground truth boxes into 2D as their input.
Our SLF supports a more practical setting that only requires regular 2D modal boxes as prompts. Thus we can generate them with an off-the-shelf detector HTC~\cite{htc} pre-trained on nuImages~\cite{nuscenes}.

\begin{table}[thb]
    \centering
    \setlength{\tabcolsep}{2pt}
    \caption{Cross dataset evaluation compared with state-of-the-art unsupervised auto-labeler FGR and supervised auto-labeler MTrans. 
    Results in gray are less informative because the NDS and error metrics are computed only on recalled samples, while FGR fails on recalling a majority of input samples (low mAP). mAOE$^\dag$ means to change the periodicity of orientation to $[0,\pi]$.}

    \begin{tabular}{l c c c c cccccc}
    \toprule
    \multirow{2}{*}{Method} & Training & \multicolumn{3}{c}{ AP$_{3D}$@KITTI } & \multicolumn{6}{c}{nuScenes Metrics} \\
        \cmidrule(lr){3-5} 
         \cmidrule(lr){6-11}
     & Data & Easy & Mod. & Hard & mAP & NDS & mATE & mASE& mAOE & mAOE$^{\dag}$ \\
    \midrule
    FGR~\cite{fgr}& - &91.3 &77.8 &70.6  &11.6 &\color{Gray}{40.4} &\color{Gray}{.152}&\color{Gray}{.188}&\color{Gray}{.010} &\color{Gray}{.010}\\
    MTrans~\cite{mtrans} & KITTI &\textbf{99.6}&\textbf{99.6}&\textbf{97.1}&  45.9 & 49.4 & \textbf{.358} & .349 & \textbf{.702} & .225 \\ 
    SLF& - &87.9&88.2&83.9& \textbf{57.2} & \textbf{55.2} & .372 & \textbf{.192} & .836 &  \textbf{.113}\\
    \bottomrule
    \end{tabular}
    
    \label{tab:supervised_results}
    
\end{table}

In~\cref{tab:supervised_results},
on KITTI, MTrans shows a nearly perfect performance,
this is not surprising since MTrans is a learnable neural network trained on  KITTI, it easily learns the specific annotation patterns in training labels.
Despite the high accuracy, we argue that learnable auto-labelers like MTrans are prone to overfit the biased pattern in the training data and generalize poorly across datasets.
When we apply MTrans to the challenging nuScenes dataset, its performance dramatically decreases.
SLF performs better in terms of mAP and NDS.
Regarding the detailed error terms, we find SLF outperforms MTrans on size estimation (mASE), and is on par with MTrans on location estimation (mATE). Regarding orientation estimation, SLF underperforms MTrans by $0.134$. However, we find a typical orientation error is the model mistakes the front end of the car for the rear end,
which is usually caused when the LiDAR points are rather sparse. 
To explore how good the orientation predictions are when the $\pi$ phase lag is disentangled, 
we change the periodicity of orientation to $[0, \pi]$ and re-measure the orientation error, denoted as mAOE$^{\dag}$. SLF outperforms MTrans on mAOE$^{\dag}$, suggesting that SLF predicts more accurate orientation, the heading direction aligning better with ground truths.
When applying FGR on the nuScences dataset, it fails on a majority of samples, resulting in a low mAP of $11.6\%$. Consequently, the NDS score and error terms are less informative since they are computed only on successfully labeled samples, typically easy samples.

\begin{figure*}
\centering
\begin{minipage}{0.48\linewidth}
\centering
\includegraphics[width=\linewidth,height=4cm]{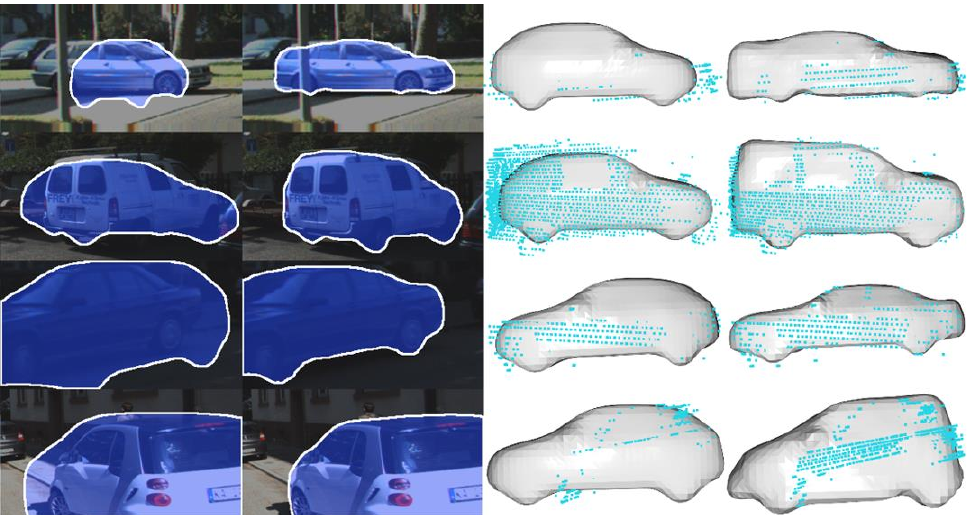}
\caption{Qualitative results of estimated shapes.
    From left to right: projected mask of the initial 3D model; projected mask of the optimized 3D model; mean shape and surrounding point cloud; optimized 3D shape and surrounding point cloud.}
    \label{fig:qualitative_results}
\end{minipage}
\hfill
\begin{minipage}{0.48\linewidth}
\centering
\includegraphics[width=\linewidth,height=4cm]{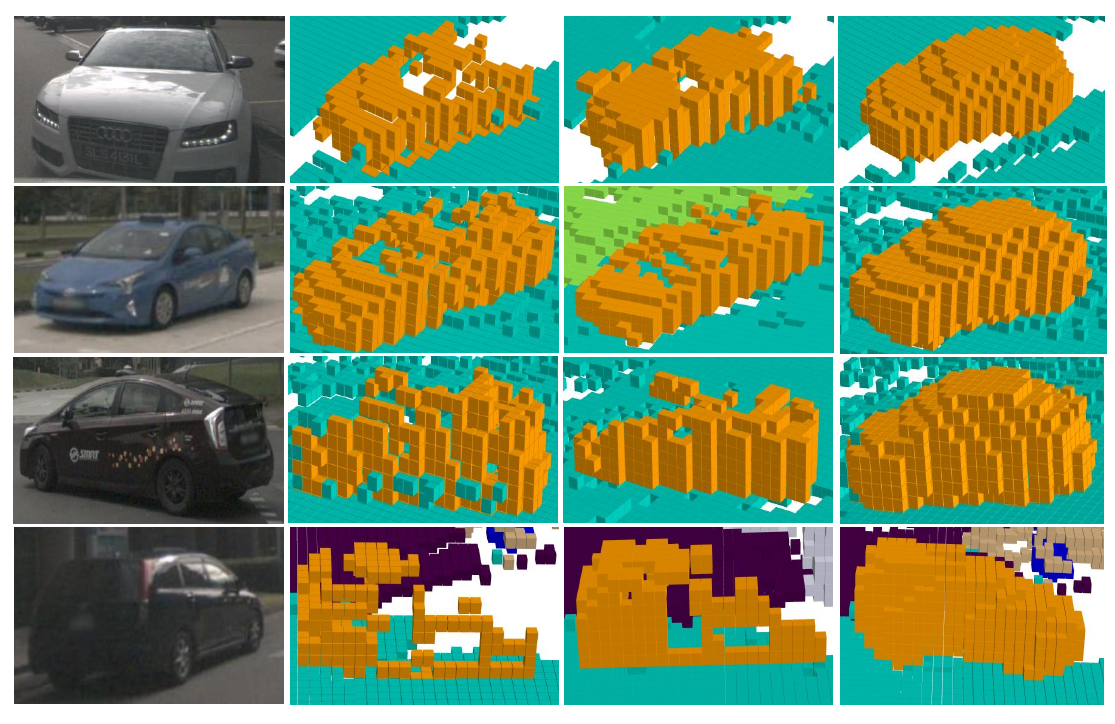}
\caption{Comparison of occupancy annotation. From left to right: reference instance image; occupancy labels from OFN~\cite{ofn}; occupancy labels from OpenOccupancy~\cite{openocc}; occupancy labels generated by our SLF.}
\label{fig:occpancy_vis}
\end{minipage}
\end{figure*}

\subsection{Voxel Occupancy Annotation}
\label{sec:shape_analysis}
Since benchmarks~\cite{kitti,nuscenes} (i.e., KITTI and nuScences)do not support quantitative evaluation on 3D shapes, we show qualitative results in \cref{fig:qualitative_results} to better show the effect of SLF on shape labeling. 
Specifically, we replace the estimated shape with a mean shape and show comparisons.  A clear improvement can be observed by comparing the initial shape and the final optimized shape, both in terms of mask alignment and LiDAR point alignment. 
Besides,
we observe that SLF can serve as a good auto-labeler for occupancy annotation of dynamic objects such as vehicles. 
We compare two versions of occupancy labels for nuScenes, OpenOccupancy~\cite{openocc} and OFN~\cite{ofn}. 
Considering that existing occupancy labels are still of low quality, quantitative evaluation may be misleading. Therefore we show qualitative visualizations in \cref{fig:occpancy_vis}, and we observe that OpenOccupancy and OFN labels fail to depict the inner occupancy of the cars. This is mainly attributed to that they label the occupancy by voxelizing the LiDAR points, and therefore self-occlusion can not be avoided. 
In contrast, SLF  can predict dense occupancy more accurately, with more detailed shapes.

\subsection{Ablation Study}
\label{sec:ablation}

In this section, we conduct comprehensive ablation studies on the KITTI validation set and report the AP$_{3D}$(IoU=0.5)$|_{R40}$ results. Except in \cref{tab:no_point}, we input 2D points to SLF, other results are obtained by using 2D bounding boxes as prompt.
\noindent\textbf{Effect of energy terms.} In \cref{tab:energy},  
we observe that introducing the mask alignment term significantly improves pseudo-label quality rather than solely applying the point cloud alignment term,
and adding the ground alignment term further improves performance.  However, as we mentioned above, using mask alignment alone can easily be trapped in sub-optimal solutions, thus it can not obtain valid results (Easy: 6.9, Mod.: 5.9, Hard: 5.6).

 \begin{figure}[thb]
\centering
    \begin{minipage}[h]{0.48\linewidth}
    
    \makeatletter\def\@captype{table}
    \setlength{\tabcolsep}{4pt}
    \caption{Ablation on energy terms.}
        \centering
        \begin{tabular}{l   c c c }
        \toprule
        Objective & Easy & Mod. & Hard \\
        \midrule
        $E_{pc}$  &  70.5 & 64.8 & 62.9 \\
        $E_{mask}$+$E_{pc}$  & 87.3 & 88.0 & 83.7 \\
        $E_{mask}$+$E_{pc}$+$E_{ground}$  &  87.9 & 88.2 & 83.9 \\
    \bottomrule
    \end{tabular}
       
       \label{tab:energy}
    \end{minipage}
    \begin{minipage}[h]{0.48\linewidth}
    
    \makeatletter\def\@captype{table}
    \setlength{\tabcolsep}{4pt}
    \caption{Ablation on 3D models.}
        \centering
        \begin{tabular}{l   c c c }
        \toprule
        \# models & Easy & Mod. & Hard \\
        \midrule
        5 & 74.7 & 70.1 & 64.0 \\ 
        30  &  84.2 & 83.6 & 77.3 \\
        79  & 87.9 & 88.2 & 83.9 \\
    \bottomrule
    \end{tabular}
        
        \label{tab:no_car}
     \end{minipage}
    
 \label{tab:ablation}
 
\end{figure}

\noindent\textbf{Effect of the number of CAD models.} In \cref{tab:no_car}, we randomly sample 5 and 30 CAD models out of the total 79 samples to investigate the influence of the number of CAD models used for PCA. From the results, we can conclude that better result comes with more models.

\noindent\textbf{Effect of the PCA dimensions.} In \cref{tab:pca_comp}, we show how different PCA dimensions affect results. 
Performance improves when we increase the dimension from 3 to 5.
More dimensions like 10 do not further improve. This is possibly attributed to that a higher dimension of shape representation may complicate the optimization process, which means it may need more steps to converge.

\noindent\textbf{Effect of the number of points.} In \cref{tab:no_point}, we analyze the effect of the number of points input to SAM~\cite{sam}. When we increase the points from 3 to 8, the improvement of pseudo-label quality is minor. However, we observe a significant decrease in the easy cases of KITTI when using one point as the prompt, which is reasonable as easy cases often have large masks and a single point can not provide sufficient context for SAM to generate high-quality results.

\begin{figure}[t]
\centering
    \begin{minipage}[h]{0.30\linewidth}
    \makeatletter\def\@captype{table}
    \setlength{\tabcolsep}{2pt}
    \caption{No. points.}

        \centering
        \begin{tabular}{l  c c c }
        \toprule
        \# Points & Easy & Mod. & Hard \\
        \midrule
        3  &  84.8 & 84.9 & 80.8 \\
        \midrule
        1 & 78.1 & 81.1 & 78.8 \\ 
        5 & 84.8 & 85.0 & 81.0\\
        8  & 85.0 & 85.2 & 80.9 \\
    \bottomrule
    \end{tabular}
    \label{tab:no_point}
     \end{minipage}
    \hfill
    \begin{minipage}[h]{0.30\linewidth}
    
    \makeatletter\def\@captype{table}
    \setlength{\tabcolsep}{2pt}
    \caption{No. dim.}

        \centering
        \begin{tabular}{l   c c c }
        \toprule
        Dim. & Easy & Mod. & Hard \\
        \midrule
        5  &  87.9 & 88.2& 83.9 \\
        \midrule
        3 & 87.1 & 87.8 & 83.6 \\ 
        8 & 86.9 & 87.1 & 84.1\\
        10  & 84.8 & 85.4 & 81.1 \\
    \bottomrule
    \end{tabular}
        
        \label{tab:pca_comp}
     \end{minipage}
    \hfill
    \begin{minipage}[h]{0.30\linewidth}
    \makeatletter\def\@captype{table}
    \setlength{\tabcolsep}{2pt}
    \caption{No. beam.}

        \centering
        \begin{tabular}{l   c c c }
        \toprule
         
        Beam & Easy & Mod. & Hard \\
        \midrule
        64 & 87.9 & 88.2 & 83.9 \\ 
        \midrule
        32  &  81.3 & 77.5 & 77.3 \\
        16  & 80.7 & 76.3 & 70.0 \\
        8 &75.1&60.0&56.7\\
    \bottomrule
    \end{tabular}
        \label{tab:beams}
     \end{minipage}

\end{figure}

\begin{figure}[t]
\centering
    \begin{minipage}[h]{0.48\linewidth}
    \makeatletter\def\@captype{table}
    \setlength{\tabcolsep}{4pt}
    \caption{Dilation on instance mask.}

        \centering
        \begin{tabular}{l   c c c }
        \toprule
        Kernel & Easy & Mod. & Hard \\
        \midrule
        FGR(no dil.) &91.3&77.8&70.6\\
        \midrule
        SLF(no dil.) & 87.9 & 88.2 & 83.9 \\ 
        5x5 &  86.3 & 85.6 & 81.2 \\
        9x9  & 83.9 & 80.5 & 76.6 \\
    \bottomrule
    \end{tabular}
       
       \label{tab:dilation}
    \end{minipage}
    \begin{minipage}[h]{0.48\linewidth}
    \makeatletter\def\@captype{table}
    \setlength{\tabcolsep}{4pt}
    \caption{Erosion on instance mask.}
        \centering
        \begin{tabular}{l   c c c }
        \toprule
        Kernel & Easy & Mod. & Hard \\
        \midrule
        FGR(no ero.) &91.3&77.8&70.6\\
        \midrule
        SLF(no ero.) & 87.9 & 88.2 & 83.9 \\ 
        5x5  &  85.2 & 83.9 & 79.5 \\
        9x9  & 82.1 & 78.2 & 72.1 \\
    \bottomrule
    \end{tabular}
        
        \label{tab:erosion}
     \end{minipage}
    
\end{figure}


\subsection{Further Analysis}
\noindent\textbf{Robustness of LiDAR Beam.} Point clouds in KITTI~\cite{kitti} are collected by 64 beams LiDAR, we downsample point cloud to 32, 16 and 8 beams following~\cite{wei2022lidar}. AP$_{3D}$ results on KITTI validation set in \cref{tab:beams} show SLF can still get decent results on 32 and 16 beams LiDAR, and the performances downgrade rapidly on Mod. and Hard cases, since these cases usually far away from ego vehicle which exacerbating the sparsity of point clouds.

\noindent\textbf{Robustness of Mask Quality.}
We simulate low-quality masks by applying erosion and dilation operations on SAM~\cite{sam} masks and report AP$_{3D}$ results of KITTI validation set in  \cref{tab:dilation} and \cref{tab:erosion}. We observe even with severely corrupted masks, SLF still outperforms the existing unsupervised auto-annotation methods in Mod. and Hard cases, which show the robustness of SLF. 

\section{Limitation}
\label{sec:lim}
The proposed SLF uses a strong shape prior learned by PCA from given 3D CAD models. The shape prior is category-specific, which means we need to prepare a collection of 3D models when labeling a specific semantic class. 

\section{Conclusion}
This paper proposes a  Segment, Lift, and Fit paradigm for automatic 3D shape annotation of objects from the 2D point or box prompts. We first \emph{segment} high-quality instance masks from the prompts using the {SAM}, and then \emph{lift} 2D masks to 3D forms, where gradient descent is used to adjust their poses and shapes until the projections \emph{fits} the masks and the surfaces conform to surrounding LiDAR points.
Extensive experimental results on multiple datasets for 3D detection demonstrate the effectiveness of our SLF method.
Moreover, the SLF auto-labeler achieves promising results in detailed shape predictions, which can be used for occupancy annotation.
In the future, we will investigate the generalization abilities by applying our SLF to more categories.


%
%
\bibliographystyle{splncs04}
\bibliography{main}
\end{document}